\documentclass[conference]{IEEEtran}
\usepackage[T1]{fontenc}
\usepackage{graphicx}
\usepackage{amsmath,amssymb,amsfonts}
\usepackage{booktabs}
\usepackage{cite}
\usepackage{url}
\IEEEoverridecommandlockouts
\begin{document}
\title{Feature-Space Latent Forecasting for Endoscopic Surgical Video Understanding}
\author{%
\IEEEauthorblockN{Songtao Hu\textsuperscript{1,*}, Hang Su\textsuperscript{1}}%
\IEEEauthorblockA{\textsuperscript{1}Xiamen University, Xiamen, China}%
\thanks{* Corresponding author: Songtao Hu (Email: \texttt{husongtao32@gmail.com}).}%
}
\maketitle
\begin{abstract}
Surgical workflow assistance is inherently anticipatory: systems must reason about how procedures evolve, not just recognize the current frame. However, many prior approaches perform prediction in pixel space by synthesizing future frames, which primarily focuses on appearance fidelity rather than the semantic representations required for downstream reasoning. To address this limitation, we propose a feature-space semantic forecasting framework that models temporal dynamics directly within the representations of a frozen Vision Foundation Model (VFM). Specifically, we build a hierarchical target representation from multi-layer VFM features, replace future-frame feature tokens with learnable \texttt{[MASK]} tokens, and train a masked transformer to forecast the corresponding future latent features from observed context, thereby formulating procedural evolution as masked future feature prediction. Experiments on public laparoscopic benchmarks demonstrate consistent improvements in phase anticipation, multi-label tool presence prediction, and remaining-time estimation over RGB-based forecasting baselines, while enabling a plug-and-play representation interface for lightweight downstream heads.
\end{abstract}

\section{Introduction}

Minimally invasive surgery videos are continuously recorded at scale in routine clinical practice, providing a rich substrate for intraoperative intelligence~\cite{twinanda2016endonet}. Yet most existing systems remain largely reactive~\cite{jin2017sv,czempiel2020tecno,liu2025lovit}, recognizing phases or tool usage only after the evidence becomes obvious~\cite{twinanda2016endonet,jin2021temporal,gao2021trans}. In practice, assistance must look ahead: it should warn of upcoming phase transitions, predict imminent tool actions, and localize where critical interactions are likely to occur, enabling timely instrument preparation and risk-aware intervention\cite{twinanda2016endonet,jin2021temporal,gao2021trans}. Therefore, a key challenge for intelligent surgical assistance is robust temporal modeling under severe appearance perturbations, so that models can learn procedural evolution from past observations and reliably forecast upcoming workflow and interactions over a future time horizon~\cite{zhang2024adaptive}.

To support intraoperative anticipation, surgical video understanding has increasingly drawn inspiration from the broader video prediction literature. In natural-video forecasting, many approaches focus on synthesizing future frames in pixel space, often through autoregressive, stochastic, or diffusion-based models~\cite{finn2016unsupervised,babaeizadeh2017stochastic,yan2021videogpt,ho2022video}. While effective for natural scenes, these methods are primarily optimized for visual fidelity. In laparoscopic surgery, however, pixel-level reconstruction is weakly aligned with workflow semantics and highly sensitive to operating-room perturbations~\cite{twinanda2016endonet,jin2021temporal,xi2025uncertainty}. Appearance fluctuations, specular highlights, smoke, and occlusions make pixel statistics unstable, so optimizing for frame synthesis may disproportionately emphasize transient photometric artifacts rather than procedure-consistent dynamics~\cite{czempiel2020tecno,liu2025lovit}. Moreover, clinically relevant evidence is often subtle and localized (e.g., near phase transitions and tool presence), while heterogeneous procedure durations and random clip sampling introduce time-scale mismatch that increases training and evaluation variance~\cite{jin2017sv,liu2023skit}. Consequently, improvements in visual realism do not necessarily translate into more consistent anticipation or robust multi-label tool prediction, revealing a gap between pixel-space objectives and downstream utility~\cite{zhang2024adaptive,gao2022simvp}.

These limitations motivate a shift from pixel-level future frame synthesis toward forecasting task-aligned representations that remain directly usable for downstream decision-making~\cite{yu2023magvit,chang2022maskgit,gupta2022maskvit}. In surgical anticipation, feature-centric pipelines already reflect this philosophy by enriching per-frame representations with instrument interactions and applying causal temporal modeling for online prediction, without relying on explicit frame decoding~\cite{yuan2022anticipation}. More broadly, recent advances in representation learning suggest that semantically structured feature spaces can provide more stable and transferable supervision for future prediction than pixel-level objectives~\cite{dosovitskiy2020image,oquab2023dinov2,amadou2024echoapex}. Vision Foundation Models (VFMs) further offer structured and transferable features that preserve consistent semantic and structural cues across visual conditions~\cite{oquab2023dinov2,radford2021learning,darcet2023vision}. This raises a central question: can VFM features serve as a semantically rich representation space for learning procedure-level dynamics and producing forecasts that downstream heads can directly consume for anticipation tasks?

Beyond radiology, pathology vision-language pre-training has also investigated learning robust representations that connect visual morphology with language supervision. Yang et al. proposed a robust baseline for pathology vision-language pre-training~\cite{yang2025enhancing}. Although histopathology images differ substantially from temporal endoscopic video, this work provides complementary evidence for studying transferable medical visual representations rather than relying solely on pixel reconstruction.

Related developments in medical image understanding further illustrate the value of representations designed for clinically meaningful outputs. Xi et al. investigated grounding radiology report findings in image segmentation, linking textual findings to spatially localized evidence~\cite{xi2026grounding}. Their MedLVR work explored latent visual reasoning for medical visual question answering~\cite{xi2026medlvr}, while a radiographic world model studied reasoning and evidence generation from medical images~\cite{xi2026radiographic}. Although these approaches address radiological images and different prediction objectives rather than endoscopic video anticipation, they motivate a complementary question: whether semantically organized latent features can also support forecasting of future surgical states without reconstructing images.

Motivated by the above, we propose a downstream-oriented framework for semantic future prediction that learns temporal dynamics directly in the feature space of a frozen Vision Foundation Model (VFM). Specifically, we construct a hierarchical target representation from multi-layer VFM features, replace the feature tokens of the future $N_p$ frames with learnable \texttt{[MASK]} tokens, and train a masked transformer to regress these masked future features conditioned on the context $N_c$ frames. This casts laparoscopic procedural evolution as feature-level forecasting, providing a direct route to semantic future prediction. By avoiding pixel-level synthesis, our objective prioritizes workflow-relevant and temporally stable evidence, such as phase boundaries, geometric structure, and deformation patterns, rather than transient textures and photometric artifacts. The forecasted features further offer a unified plug-and-play interface that can be consumed by lightweight downstream heads for phase anticipation and multi-label tool presence prediction, enabling efficient adaptation with limited supervision and improved transfer across tasks and datasets.

\begin{figure*}[t]
  \centering
  \includegraphics[width=\textwidth]{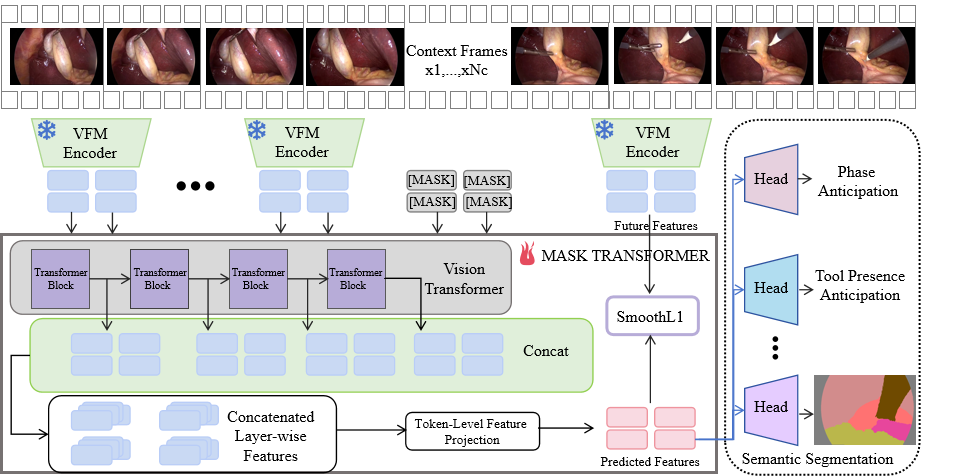}
  \caption{Overview of our feature-space forecasting pipeline for endoscopic video understanding. A frozen VFM encoder extracts multi-layer hierarchical features from observed context frames, while future-frame tokens are replaced by learnable \texttt{[MASK]} placeholders. A masked vision transformer predicts compact future feature maps, which are then fed into plug-and-play task heads.}
  \label{fig:overview}
\end{figure*}

\section{Method}

\subsection{Overview}
Given a laparoscopic surgery video $X=\{x_t\}_{t=1}^{T}$, we sample a subsequence of length $N=N_c+N_p$ to form a forecasting clip, where the first $N_c$ frames serve as observed context and the remaining $N_p$ frames define the prediction horizon. Our objective is to forecast future representations from the observed context and ensure that these forecasts can be directly consumed by downstream surgical understanding tasks such as workflow anticipation and tool usage modeling. The framework comprises three components: a frozen feature extractor $\Phi$ that produces multi-level frame representations, a trainable predictor $\Psi_\theta$ that infers future features conditioned on the observed context, and a plug-and-play task head $g(\cdot)$ operating on the predicted features. We keep $\Phi$ fixed throughout training and optimize $\Psi_\theta$ (and optionally $g$), thereby decoupling temporal forecasting from representation learning and reducing sensitivity to low-level appearance variations. 

\subsection{Hierarchical Target Representation}
As depth increases in a Vision Transformer, patch tokens progressively assume a proxy role by aggregating information from neighboring regions through self-attention, forming higher-level structural units\cite{bolya2025perception}. At the same time, the CLS token continuously integrates interactions across all patch tokens and gradually evolves into a compact representation of the global semantic state of the frame\cite{wu2023clipself}. This depth-wise structural evolution from local patch representations to intermediate semantic aggregation and ultimately to global summarization indicates that different layers encode complementary levels of structural abstraction. Relying on a single layer would therefore restrict forecasting to a limited abstraction scale. To jointly capture fine-grained tool–tissue interactions and higher-level procedural dynamics, we concatenate token representations from multiple depths, constructing a unified feature space for temporal prediction.

Let $\Phi$ denote a pretrained ViT-based encoder kept frozen throughout training. Given a frame $x_t$, we extract intermediate features from a selected set of layers $\mathcal{L}$:
\begin{equation}
f_t^{(\ell)} = \Phi^{(\ell)}(x_t), \qquad \ell \in \mathcal{L},
\end{equation}
where $f_t^{(\ell)} \in \mathbb{R}^{H\times W\times D_\ell}$ is the patch-grid feature map at layer $\ell$. Because different layers capture complementary information, we fuse them by concatenating along the channel dimension,
\begin{equation}
f_t^{\text{cat}} = \text{Concat}_{\ell\in\mathcal{L}} f_t^{(\ell)} \in \mathbb{R}^{H\times W\times D_{\text{cat}}},
\end{equation}
where $D_{\text{cat}}=\sum_{\ell\in\mathcal{L}} D_\ell$. Directly forecasting $f_t^{\text{cat}}$ can be unnecessarily difficult due to its high dimensionality and redundancy. We thus apply a fixed dimensionality reduction $\Pi(\cdot)$ to obtain a compact target space:
\begin{equation}
z_t = \Pi\!\left(f_t^{\text{cat}}\right), \qquad z_t \in \mathbb{R}^{H\times W\times D}, \ \ D \ll D_{\text{cat}}.
\end{equation}
In practice, $\Pi$ is implemented as an offline PCA projection fitted on the training set. The projection is then held fixed so that learning focuses on temporal forecasting in a stable target space rather than reshaping the representation itself \cite{karypidis2024dino}. The resulting compact features $\{z_t\}$ serve as our prediction targets for future-frame forecasting and form a unified interface that can be consumed by lightweight downstream heads.

\subsection{Future Feature Forecasting in Feature Space}
Given a forecasting clip of length $N=N_c+N_p$, we perform prediction directly in the compact target feature space.
For each frame, we reshape the target features $z_t\in\mathbb{R}^{H\times W\times D}$ into a token sequence
\begin{equation}
s_t=\mathrm{Flatten}(z_t)\in\mathbb{R}^{(HW)\times D},
\end{equation}
where $\mathrm{Flatten}(\cdot)$ follows a fixed raster-scan order over the $H\times W$ grid and each token corresponds to a unique spatial location.
To retain both temporal ordering and spatial layout, we add temporal and spatial positional embeddings to each token:
\begin{equation}
u_t(i) = s_t(i) + e_t^{\mathrm{temp}} + e^{\mathrm{spat}}(i), \qquad i\in\{1,\dots,HW\},
\label{eq:pos_embed}
\end{equation}
where $e_t^{\mathrm{temp}}\in\mathbb{R}^{D}$ is a learnable temporal embedding for frame $t$ and
$e^{\mathrm{spat}}(i)\in\mathbb{R}^{D}$ is a learnable spatial embedding for token index $i$ (broadcast to match the token dimension).

\paragraph{Future slots as learnable placeholders.}
A core design is the explicit parameterization of the future segment.
For observed frames ($t\le N_c$), we use the true tokens $u_t$.
For future frames ($t>N_c$), we replace the (unknown) content tokens with a shared learnable placeholder vector $m\in\mathbb{R}^{D}$,
which is shared across all future time steps and all spatial locations, while keeping the corresponding positional information:
\begin{equation}
\tilde u_t(i)=
\begin{cases}
u_t(i), & t\le N_c,\\[2pt]
m + e_t^{\mathrm{temp}} + e^{\mathrm{spat}}(i), & t> N_c,
\end{cases}
\label{eq:future_placeholder}
\end{equation}
This construction marks which time steps are to be predicted and preserves token-wise spatial correspondence,
while forcing the model to infer future content from the observed context.

\paragraph{Masked transformer predictor with factorized attention.}
We concatenate tokens across time as $\tilde U=[\tilde u_1;\dots;\tilde u_N]\in\mathbb{R}^{(N\cdot HW)\times D}$ (ordered first by time, then by spatial index)
and feed them into a transformer predictor
\begin{equation}
\hat U=\Psi_\theta(\tilde U).
\end{equation}
We use full bidirectional attention over $\tilde U$ and apply supervision only on the future positions ($t>N_c$).
The output $\hat U$ is reshaped back to per-frame feature maps, yielding predicted features $\hat z_t\in\mathbb{R}^{H\times W\times D}$ for $t>N_c$.
To handle high-resolution grids efficiently, each transformer block factorizes attention into two steps:
(i) \emph{temporal attention} is applied within groups $\{\tilde u_1(i),\dots,\tilde u_N(i)\}$ at fixed spatial index $i$,
and (ii) \emph{spatial attention} is applied within each frame over $\{\tilde u_t(1),\dots,\tilde u_t(HW)\}$.
Stacking such blocks propagates information across time while preserving within-frame structural coherence and improves computational efficiency for high-resolution grids.

\paragraph{Training objective.}
We train $\Psi_\theta$ by regressing only the future features using a robust SmoothL1 loss applied element-wise over the channel dimension and averaged over tokens:
\begin{equation}
\mathcal{L}_{\mathrm{pred}}
=
\frac{1}{N_pHW}
\sum_{t=N_c+1}^{N}\sum_{i=1}^{HW}
\mathrm{SmoothL1}\!\left(\hat z_t(i),\, z_t(i)\right),
\label{eq:pred_loss}
\end{equation}
where $z_t(i)\in\mathbb{R}^{D}$ denotes the target feature vector at token index $i$ (equivalently location $(u,v)$ on the $H\times W$ grid).
At inference time, we provide only the observed frames, fill future slots using Eq.~\eqref{eq:future_placeholder},
and obtain $\{\hat z_t\}_{t=N_c+1}^{N}$.
The forecasted features form a unified interface for downstream heads, e.g.,
$\hat y = g_{\mathrm{phase}}(\{\hat z_t\})$ for phase anticipation and $\hat{\mathbf{p}} = g_{\mathrm{tool}}(\{\hat z_t\})$ for multi-label tool presence prediction,
enabling modular reuse across tasks and datasets.

\subsection{Forecasting Interface and Information Flow}
\label{sec:info_flow}
The proposed formulation separates the information available at prediction time from the information used as supervision. During training, future images are passed through the frozen encoder only to construct the target tensors $z_{N_c+1:N}$; they do not enter the input sequence of the predictor. During inference, the same future positions are occupied entirely by learned placeholders. This distinction is important for anticipation: allowing any future-frame image content to enter the predictor would turn forecasting into feature reconstruction with privileged observations and invalidate the causal evaluation. Although attention inside the predictor is bidirectional across the concatenated token sequence, the only content-bearing tokens come from the observed context. Bidirectionality therefore allows future placeholder positions to interact without exposing future visual evidence.

The representation $z_t$ also preserves a spatial grid rather than collapsing every frame into a single embedding. This design provides two complementary pathways for surgical reasoning. Temporal attention can propagate motion and progression evidence across the observed clip, whereas spatial attention can combine cues from multiple locations, including tool regions and surrounding tissue. The same predicted tensor can subsequently support global classification through pooling or spatial prediction through a dense head. This is a property of the representation interface rather than evidence that every downstream task benefits equally; its empirical usefulness is evaluated separately for each task.

\subsection{Evaluation Controls and Interpretation}
\label{sec:eval_controls}
A meaningful anticipation comparison requires matching the time point being predicted. In our experiments, the input context ends at $t$, while the target label refers to $t+\Delta$. The Copy-Last baseline supplies a useful counterfactual: it assumes that the last observed representation remains adequate at the future target time. Improvements over this baseline therefore indicate that the learned forecast contains information beyond feature persistence. Conversely, Oracle uses the encoder feature extracted from the actual future image and quantifies the performance possible for a downstream head when future visual evidence is available. Oracle is not a deployable anticipatory model and should not be interpreted as a fair online competitor.

These controls also clarify the scope of the RGB comparison. An RGB generator must first synthesize an image and then convert the generated appearance into a representation usable by a task head. Our method forecasts the representation directly, eliminating the image-decoding stage from the inference pathway. The experimental comparison tests the downstream utility of these two routes, not whether feature forecasting can produce more photorealistic future frames. In addition, the current-frame segmentation rows in Table~\ref{tab:current_vs_future} are included as reference points for conventional recognition and are intentionally separated from future-frame anticipation results. Pooling the two settings into a single ranking would conflate access to current visual evidence with prediction of unobserved evidence.

\section{Experiments}
\subsection{Dataset}
\label{sec:exp_dataset}
We evaluate intraoperative anticipation on Cholec80, a widely used benchmark consisting of 80 laparoscopic cholecystectomy videos with dense frame-level workflow annotations. Each frame is labeled with (i) a single surgical phase from seven classes (Preparation, CalotTriangleDissection, ClippingCutting, GallbladderDissection, GallbladderPackaging, CleaningCoagulation, GallbladderRetraction) and (ii) a multi-label tool-presence vector over seven instruments (Grasper, Bipolar, Hook, Scissors, Clipper, Irrigator, SpecimenBag). These two annotation streams support both coarse-grained workflow modeling (phase) and fine-grained procedural context (tools), and remaining-time-to-transition targets are deterministically derived from the phase timeline. We adopt the Cholec80 train/val/test split and report results on the held-out test set unless otherwise stated~\cite{yuan2022anticipation}.
Segmentation experiments follow the standard protocol~\cite{silva2022analysis}.

\begin{table*}[!t]
\caption{\textbf{Cholec80 phase recognition.} Forecasting future features improves macro-F1 over the RGB baseline; we report Acc., mF1, and per-phase F1.}
\label{tab:cholec80_phase_test}
\centering
\footnotesize
\setlength{\tabcolsep}{3.2pt}
\renewcommand{\arraystretch}{1.08}
\resizebox{\textwidth}{!}{%
\begin{tabular}{l|cc|ccccccc}
\hline
\textbf{Method} & \textbf{Acc.} & \textbf{mF1} &
\textbf{Prep} & \textbf{Calot} & \textbf{C\&C} & \textbf{GDis} &
\textbf{GPack} & \textbf{Clean} & \textbf{GRet} \\
\hline
Baseline (RGB)        & 0.5118 & 0.3626 & 0.2633 & 0.5552 & 0.0115 & 0.6562 & 0.2847 & 0.4540 & 0.3132 \\
Oracle (GT feat.)     & 0.7006 & 0.6570 & 0.6369 & 0.6022 & 0.3135 & 0.8879 & 0.6800 & 0.8292 & 0.6490 \\
Copy-Last (feat.)     & 0.6547 & 0.5799 & 0.5586 & 0.5834 & 0.1812 & 0.8423 & 0.5987 & 0.7618 & 0.5334 \\
Ours (forecast feat.) & 0.6753 & 0.6101 & 0.6016 & 0.6062 & 0.2314 & 0.8617 & 0.6389 & 0.7896 & 0.5414 \\
\hline
\end{tabular}%
}
\vspace{-2mm}
\end{table*}

\begin{table}[!t]
\caption{\textbf{Cholec80 tool presence recognition.} Macro-F1 (mF1) and per-tool F1 comparing the RGB baseline, feature-space baselines, and our forecast features.}
\label{tab:cholec80_tool_test}
\centering
\footnotesize
\setlength{\tabcolsep}{3.6pt}
\renewcommand{\arraystretch}{1.08}
\resizebox{\linewidth}{!}{%
\begin{tabular}{l|c|ccccccc}
\hline
\textbf{Method} & \textbf{mF1} &
\textbf{Grasp} & \textbf{Bip} & \textbf{Hook} & \textbf{Scis} &
\textbf{Clip} & \textbf{Irrig} & \textbf{Bag} \\
\hline
Baseline (RGB)        & 0.4145 & 0.6631 & 0.2974 & 0.8046 & 0.0817 & 0.1789 & 0.3572 & 0.5186 \\
Oracle (GT feat.)     & 0.5756 & 0.6858 & 0.7394 & 0.9258 & 0.2555 & 0.5970 & 0.2893 & 0.5362 \\
Copy-Last (feat.)     & 0.5267 & 0.6486 & 0.6591 & 0.8709 & 0.2467 & 0.5433 & 0.2447 & 0.4736 \\
Ours (forecast feat.) & 0.5521 & 0.6656 & 0.6862 & 0.9087 & 0.2503 & 0.5635 & 0.2690 & 0.5213 \\
\hline
\end{tabular}%
}
\vspace{-2mm}
\end{table}

\subsection{Implementation Details}
\label{sec:exp_impl}
Our pipeline disentangles temporal modeling from downstream head capacity by extracting frozen per-frame features $F_t$, training a lightweight forecaster to predict $\hat{F}_{t+\Delta}$ with $\Delta=1$\,s from an observed context window, and evaluating task-specific heads on the representation used at test time. We compare three feature inputs while keeping the head architecture and its training protocol identical: Copy-Last uses the last observed feature $F_t$ as a persistence baseline, Ours feeds the forecasted feature $\hat{F}_{t+\Delta}$, and Oracle provides the ground-truth future feature $F_{t+\Delta}$ as an upper bound. This design ensures that performance differences reflect representation quality rather than classifier complexity. Experiments are run on two NVIDIA RTX PRO 6000 GPUs (96\,GB each) using AdamW with a cosine learning-rate schedule and automatic mixed precision.

\subsection{Downstream Tasks and Evaluation Protocol}
\label{sec:exp_tasks}

\paragraph{(1) Phase anticipation.}
We predict the surgical phase at a future time $t{+}\Delta$, denoted by $y^{\text{phase}}_{t+\Delta}\in\{1,\dots,7\}$. Phase anticipation is practically important because assistance must act before transitions become visually obvious, enabling timely preparation around imminent workflow changes. As an RGB baseline, we train a pixel-space forecaster based on SimVP to predict the next RGB frame from the observed context, and then apply the same phase head at $t{+}\Delta$ on the predicted frame~\cite{gao2022simvp}. Table~\ref{tab:cholec80_phase_test} reports Acc.\ and macro-F1 (mF1). Compared with the RGB baseline (Acc.\ 0.5118, mF1 0.3626), our forecast features achieve 0.6753 Acc.\ and 0.6101 mF1, improving by +0.1635 Acc.\ and +0.2475 mF1, while consistently strengthening per-phase F1, including transition-heavy classes such as ClippingCutting (0.0115$\rightarrow$0.2314).

\paragraph{(2) Tool presence anticipation.}
We predict a future multi-label tool vector $\mathbf{y}^{\text{tool}}_{t+\Delta}\in\{0,1\}^{7}$ indicating the presence of each instrument at time $t{+}\Delta$. Tool anticipation provides a fine-grained cue about imminent actions and interactions, complementing phase anticipation for decision-critical assistance. Table~\ref{tab:cholec80_tool_test} reports macro-F1 (mF1) and per-tool F1. Our forecast features improve mF1 from 0.4145 (RGB baseline) to 0.5521 (+0.1376), with gains that are not restricted to a single frequent instrument and include large improvements on context-sensitive tools such as Bipolar (0.2974$\rightarrow$0.6862) and Clipper (0.1789$\rightarrow$0.5635).

\begin{table*}[!t]
\caption{\textbf{Remaining time to next phase occurrence on Cholec80 (60--20).} Mean absolute error (MAE, minutes) at 2/3/5-minute horizons, reported for samples \emph{inside} and \emph{outside} the anticipation window, together with their weighted mean.}
\label{tab:cholec80_rtt_mae}
\centering
\footnotesize
\setlength{\tabcolsep}{4.2pt}
\renewcommand{\arraystretch}{0.8}
\resizebox{\textwidth}{!}{%
\begin{tabular}{l|ccc|ccc|ccc}
\hline
\textbf{Method} & \multicolumn{3}{c|}{\textbf{wMAE} $\downarrow$} & \multicolumn{3}{c|}{\textbf{inMAE} $\downarrow$} & \multicolumn{3}{c}{\textbf{outMAE} $\downarrow$} \\
 & \textbf{2m} & \textbf{3m} & \textbf{5m} & \textbf{2m} & \textbf{3m} & \textbf{5m} & \textbf{2m} & \textbf{3m} & \textbf{5m} \\
\hline
Bayesian  & 0.39 & 0.59 & 0.85 & 0.63 & 0.86 & 1.17 & 0.15 & 0.32 & 0.52 \\
IIA-Net                       & 0.36 & 0.49 & 0.68 & 0.62 & 0.81 & 1.08 & 0.10 & 0.18 & 0.28 \\
Persistence (Copy-Last)                    & 0.35 & 0.51 & 0.73 & 0.61 & 0.80 & 1.10 & 0.10 & 0.17 & 0.29 \\
Ours (Forecast)               & 0.34 & 0.48 & 0.66 & 0.59 & 0.78 & 1.06 & 0.09 & 0.16 & 0.26 \\
Oracle                   & 0.33 & 0.47 & 0.64 & 0.58 & 0.76 & 1.04 & 0.09 & 0.15 & 0.24 \\
\hline
\end{tabular}%
}
\vspace{-2mm}
\end{table*}

\paragraph{(3) Remaining time to next phase (RTT).}
Given the context up to time $t$, the model predicts the remaining time until the next phase transition. We report mean absolute error (MAE, minutes) at 2/3/5-minute horizons, decomposed into inMAE for samples inside the anticipation window, outMAE for samples outside the window, and their weighted mean (wMAE). This breakdown is diagnostic because in-window samples are closer to transitions and often more ambiguous, while out-of-window samples reflect longer-range procedural progression where cues can be subtle and style-dependent. We include a Bayesian baseline for RTT prediction.~\cite{rivoir2020rethinking}
We also compare against an interaction-aware anticipation model (IIA-Net) as a strong prior method.~\cite{yuan2022anticipation}
As shown in Table~\ref{tab:cholec80_rtt_mae}, our forecast features consistently reduce wMAE compared with persistence across all horizons, from 0.35/0.51/0.73 to 0.34/0.48/0.66 at 2/3/5 minutes, approaching the oracle upper bound (0.33/0.47/0.64).

\begin{table*}[!t]
\caption{\textbf{Quantitative results under two evaluation settings.}
\emph{(A) is current-frame segmentation; (B) is future anticipation at $t{+}\Delta$ and is not directly comparable to (A).}}
\label{tab:current_vs_future}
\centering
\footnotesize
\setlength{\tabcolsep}{3.8pt}
\renewcommand{\arraystretch}{1.40}
\resizebox{\textwidth}{!}{%
\begin{tabular}{l|cccccccc|cccc}
\hline
\multicolumn{13}{l}{\textbf{(A) Current-frame segmentation baselines (frame $t$)}}\\
\hline
\textbf{Networks} &
\textbf{Dice Bkg} & \textbf{Dice AW} & \textbf{Dice Liver} & \textbf{Dice GT} &
\textbf{Dice Fat} & \textbf{Dice Ins} & \textbf{Dice Gb} & \textbf{Dice Misc.} &
\textbf{Dice Mean} & \textbf{IoU Mean} & \textbf{Recall Mean} & \textbf{Precision Mean} \\
\hline
U-Net        & 0.97 & 0.73 & 0.52 & 0.26 & 0.87 & 0.52 & 0.40 & 0.08 & 0.54 & 0.43 & 0.67 & 0.55 \\
U-Net++      & 0.98 & 0.83 & 0.75 & 0.11 & 0.91 & 0.61 & 0.63 & 0.16 & 0.62 & 0.55 & 0.72 & 0.73 \\
DyUNet       & 0.98 & 0.84 & 0.82 & 0.13 & 0.89 & 0.57 & 0.57 & 0.00 & 0.60 & 0.52 & 0.70 & 0.63 \\
UNETR        & 0.97 & 0.76 & 0.74 & 0.35 & 0.88 & 0.71 & 0.42 & 0.00 & 0.60 & 0.49 & 0.72 & 0.59 \\
DeepLabV3+   & 0.97 & 0.81 & 0.74 & 0.14 & 0.86 & 0.62 & 0.60 & 0.12 & 0.61 & 0.50 & 0.69 & 0.62 \\
\hline
\multicolumn{13}{l}{\textbf{(B) Future anticipation setting (frame $t{+}\Delta$)}}\\
\hline
\textbf{Methods} &
\textbf{Dice Bkg} & \textbf{Dice AW} & \textbf{Dice Liver} & \textbf{Dice GT} &
\textbf{Dice Fat} & \textbf{Dice Ins} & \textbf{Dice Gb} & \textbf{Dice Misc.} &
\textbf{Dice Mean} & \textbf{IoU Mean} & \textbf{Recall Mean} & \textbf{Precision Mean} \\
\hline
VideoCrafter2 (RGB diffusion$\rightarrow$feat) 
& 0.95 & 0.68 & 0.47 & 0.08 & 0.74 & 0.49 & 0.35 & 0.01
& 0.47 & 0.36 & 0.50 & 0.44 \\
Copy-Last    & 0.97 & 0.79 & 0.70 & 0.14 & 0.88 & 0.57 & 0.54 & 0.08 & 0.59 & 0.47 & 0.69 & 0.64 \\
Ours         & 0.97 & 0.81 & 0.72 & 0.17 & 0.89 & 0.60 & 0.57 & 0.10 & 0.61 & 0.49 & 0.71 & 0.64 \\
Oracle (GT)  & 0.98 & 0.82 & 0.74 & 0.19 & 0.90 & 0.62 & 0.59 & 0.11 & 0.62 & 0.50 & 0.72 & 0.65 \\

\hline
\end{tabular}%
}
\vspace{-2mm}
\end{table*}

\paragraph{(4) Controlled comparison under current-frame vs future settings.}
We report results under two evaluation settings that serve different purposes. In (A), we benchmark standard segmentation networks on the current frame $t$ when full visual evidence is available at test time. In (B), we evaluate segmentation anticipation at $t{+}\Delta$ using a fixed segmentation head and varying only the representation source to isolate the effect of temporal modeling, including a VideoCrafter2 latent-diffusion baseline that generates future RGB frames before feature extraction.

\begin{figure}[t]
    \centering
    \includegraphics[width=0.8\linewidth]{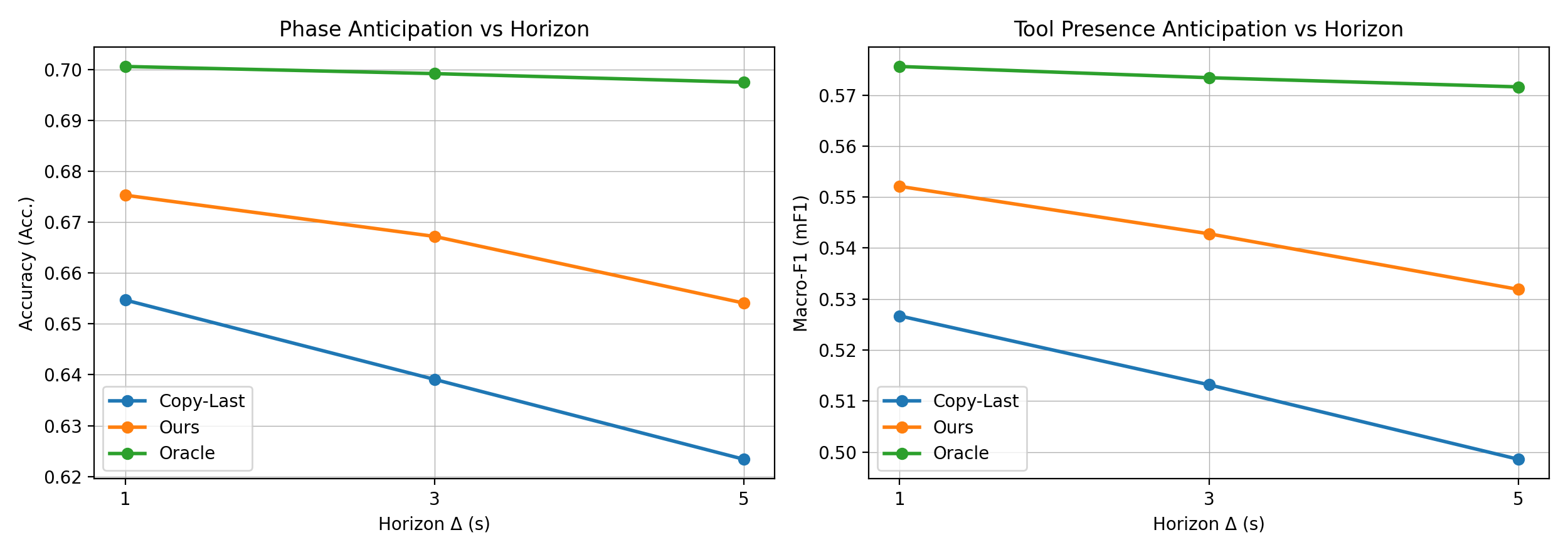}
    \caption{\textbf{Horizon analysis on Cholec80.} Performance as a function of anticipation horizon $\Delta\in\{1,3,5\}$ seconds. We keep the downstream head and protocol fixed and vary only the representation source (Copy-Last, Ours, Oracle).}
    \label{fig:horizon}
\end{figure}

\paragraph{(5) Horizon sensitivity.}
As shown in Fig.~\ref{fig:horizon}, We evaluate robustness across horizons $\Delta\in{1,3,5}$s under a controlled setting (fixed head; varying only feature source). While performance degrades mildly with larger $\Delta$, our forecasts consistently outperform Copy-Last and stay closer to the oracle upper bound. Notably, even at $\Delta=5$s, they exceed the RGB baseline at $\Delta=1$s, highlighting the effectiveness of VFM feature-space forecasting for longer-horizon anticipation.

\subsection{Further Analysis of the Reported Results}
\label{sec:result_analysis}
\paragraph{Temporal forecasting versus persistence.}
The phase anticipation comparison in Table~\ref{tab:cholec80_phase_test} shows that Ours improves macro-F1 from 0.5799 to 0.6101 relative to Copy-Last, an absolute gain of 0.0302. Accuracy rises from 0.6547 to 0.6753. For tool presence, the corresponding macro-F1 values are 0.5267 and 0.5521, giving a gain of 0.0254. These comparisons are more targeted than the RGB baseline comparison because they hold the feature representation family fixed and isolate the contribution of future-state estimation. The improvements are modest relative to the gap between the RGB baseline and the feature-based methods, suggesting that both the choice of representation and the temporal predictor contribute to final performance.

\paragraph{Class-wise behavior and remaining gaps.}
The phase-level scores expose differences masked by aggregate metrics. For ClippingCutting, Ours raises F1 from 0.1812 (Copy-Last) to 0.2314, but remains below the Oracle score of 0.3135. GallbladderDissection is comparatively strong (0.8617 versus an Oracle value of 0.8879), while GallbladderRetraction still shows a noticeable gap (0.5414 versus 0.6490). Tool classes display similar variability: Hook reaches 0.9087 F1, close to the Oracle value of 0.9258, whereas Scissors remains at 0.2503 and Irrigator at 0.2690. These reported values indicate that predicting a useful general workflow state does not guarantee equally reliable evidence for brief or visually ambiguous events. They also caution against using macro-F1 alone to characterize the practical reliability of every instrument category.

\paragraph{Remaining-time errors.}
The RTT table reveals that the forecasting advantage persists across the reported anticipation windows. Relative to Copy-Last, weighted MAE is lower by 0.01, 0.03, and 0.07 minutes at the 2-, 3-, and 5-minute horizons, respectively. The larger absolute difference at 5 minutes is consistent with the intuition that a static representation loses relevance as the future target becomes more distant. Nevertheless, errors are not eliminated, and all forecasting results remain above the corresponding Oracle values. Moreover, the in-window MAE is considerably greater than out-of-window MAE for all methods, showing that accurate transition timing remains difficult near the decision boundary. Since these results are descriptive, they should not be interpreted as evidence of statistical significance without confidence intervals or repeated-run variability.

\paragraph{Dense prediction as a representation probe.}
For the future segmentation setting in Table~\ref{tab:current_vs_future}, Ours achieves mean Dice of 0.61 compared with 0.59 for Copy-Last and 0.47 for the VideoCrafter2-based RGB route; the Oracle future feature yields 0.62. This result supports the use of a spatially organized latent representation, but should be interpreted in the context of the stated evaluation protocol. In particular, it is not a comparison between the anticipation model and the current-frame segmentation architectures in part (A) of the table. Per-class values further show that improvements for smaller structures remain limited: the GT and miscellaneous classes attain Dice scores of 0.17 and 0.10, respectively, under forecasting. Thus, aggregate overlap metrics should be considered alongside the weaker categories when judging suitability for applications that depend on small anatomical or instrument regions.

\subsection{Limitations and Future Directions}
\label{sec:limitations}
Our framework inherits the coverage and biases of its frozen visual encoder. Features pretrained outside a particular surgical setting may fail to preserve uncommon instruments, subtle tissue changes, or cues obscured by smoke and bleeding. Furthermore, PCA deliberately compresses multi-layer features, so information that is low-variance in the training distribution can be attenuated even when it matters for a rare clinical event. Testing alternative compression strategies or spatially selective losses would help determine whether such cues can be retained without a large increase in forecasting cost.

The reported experiments emphasize short-term anticipation on laparoscopic benchmarks. They do not establish generalization to other hospitals, procedures, imaging systems, or substantial shifts in workflow. Forecasts may also become less reliable during abrupt tool entry, unexpected intervention, and phases with variable duration, where a deterministic latent regression objective may average over several plausible futures. Probabilistic feature forecasting and explicit uncertainty estimation are natural next steps, especially when a model is used to trigger warnings. Finally, predictive performance on annotated videos is not itself evidence of clinical safety or improved outcomes. Any prospective deployment would require external validation, calibrated confidence estimates, latency measurements, and human oversight.

\section{Conclusion}
We propose a reconstruction-free feature forecasting framework that predicts future states directly in frozen VFM feature space via masked completion, avoiding pixel synthesis and focusing on semantically stable procedural dynamics. On Cholec80, it consistently improves phase, tool, and remaining-time anticipation toward the oracle upper bound, while remaining training-efficient with a frozen backbone and offering a plug-and-play interface for lightweight task heads across tasks and evaluation protocols.

\bibliographystyle{IEEEtran}
\bibliography{refs}

@article{babaeizadeh2017stochastic,
  title={Stochastic variational video prediction},
  author={Babaeizadeh, Mohammad and Finn, Chelsea and Erhan, Dumitru and Campbell, Roy H and Levine, Sergey},
  journal={arXiv preprint arXiv:1710.11252},
  year={2017}
}

@article{finn2016unsupervised,
  title={Unsupervised learning for physical interaction through video prediction},
  author={Finn, Chelsea and Goodfellow, Ian and Levine, Sergey},
  journal={Advances in neural information processing systems},
  volume={29},
  year={2016}
}

@article{ho2022video,
  title={Video diffusion models},
  author={Ho, Jonathan and Salimans, Tim and Gritsenko, Alexey and Chan, William and Norouzi, Mohammad and Fleet, David J},
  journal={Advances in neural information processing systems},
  volume={35},
  pages={8633--8646},
  year={2022}
}

@inproceedings{gao2022simvp,
  title={Simvp: Simpler yet better video prediction},
  author={Gao, Zhangyang and Tan, Cheng and Wu, Lirong and Li, Stan Z},
  booktitle={Proceedings of the IEEE/CVF conference on computer vision and pattern recognition},
  pages={3170--3180},
  year={2022}
}

@article{yan2021videogpt,
  title={Videogpt: Video generation using vq-vae and transformers},
  author={Yan, Wilson and Zhang, Yunzhi and Abbeel, Pieter and Srinivas, Aravind},
  journal={arXiv preprint arXiv:2104.10157},
  year={2021}
}

@inproceedings{yu2023magvit,
  title={Magvit: Masked generative video transformer},
  author={Yu, Lijun and Cheng, Yong and Sohn, Kihyuk and Lezama, Jos{\'e} and Zhang, Han and Chang, Huiwen and Hauptmann, Alexander G and Yang, Ming-Hsuan and Hao, Yuan and Essa, Irfan and others},
  booktitle={Proceedings of the IEEE/CVF Conference on Computer Vision and Pattern Recognition},
  pages={10459--10469},
  year={2023}
}

@inproceedings{chang2022maskgit,
  title={Maskgit: Masked generative image transformer},
  author={Chang, Huiwen and Zhang, Han and Jiang, Lu and Liu, Ce and Freeman, William T},
  booktitle={Proceedings of the IEEE/CVF conference on computer vision and pattern recognition},
  pages={11315--11325},
  year={2022}
}

@article{oquab2023dinov2,
  title={Dinov2: Learning robust visual features without supervision},
  author={Oquab, Maxime and Darcet, Timoth{\'e}e and Moutakanni, Th{\'e}o and Vo, Huy and Szafraniec, Marc and Khalidov, Vasil and Fernandez, Pierre and Haziza, Daniel and Massa, Francisco and El-Nouby, Alaaeldin and others},
  journal={arXiv preprint arXiv:2304.07193},
  year={2023}
}

@inproceedings{radford2021learning,
  title={Learning transferable visual models from natural language supervision},
  author={Radford, Alec and Kim, Jong Wook and Hallacy, Chris and Ramesh, Aditya and Goh, Gabriel and Agarwal, Sandhini and Sastry, Girish and Askell, Amanda and Mishkin, Pamela and Clark, Jack and others},
  booktitle={International conference on machine learning},
  pages={8748--8763},
  year={2021},
  organization={PmLR}
}

@article{darcet2023vision,
  title={Vision transformers need registers},
  author={Darcet, Timoth{\'e}e and Oquab, Maxime and Mairal, Julien and Bojanowski, Piotr},
  journal={arXiv preprint arXiv:2309.16588},
  year={2023}
}

@article{dosovitskiy2020image,
  title={An image is worth 16x16 words: Transformers for image recognition at scale},
  author={Dosovitskiy, Alexey},
  journal={arXiv preprint arXiv:2010.11929},
  year={2020}
}

@article{gupta2022maskvit,
  title={Maskvit: Masked visual pre-training for video prediction},
  author={Gupta, Agrim and Tian, Stephen and Zhang, Yunzhi and Wu, Jiajun and Mart{\'\i}n-Mart{\'\i}n, Roberto and Fei-Fei, Li},
  journal={arXiv preprint arXiv:2206.11894},
  year={2022}
}

@article{amadou2024echoapex,
  title={Echoapex: A general-purpose vision foundation model for echocardiography},
  author={Amadou, Abdoul Aziz and Zhang, Yue and Piat, Sebastien and Klein, Paul and Schmuecking, Ingo and Passerini, Tiziano and Sharma, Puneet},
  journal={arXiv preprint arXiv:2410.11092},
  year={2024}
}

@article{twinanda2016endonet,
  title={Endonet: a deep architecture for recognition tasks on laparoscopic videos},
  author={Twinanda, Andru P and Shehata, Sherif and Mutter, Didier and Marescaux, Jacques and De Mathelin, Michel and Padoy, Nicolas},
  journal={IEEE transactions on medical imaging},
  volume={36},
  number={1},
  pages={86--97},
  year={2016},
  publisher={IEEE}
}

@article{jin2017sv,
  title={SV-RCNet: workflow recognition from surgical videos using recurrent convolutional network},
  author={Jin, Yueming and Dou, Qi and Chen, Hao and Yu, Lequan and Qin, Jing and Fu, Chi-Wing and Heng, Pheng-Ann},
  journal={IEEE transactions on medical imaging},
  volume={37},
  number={5},
  pages={1114--1126},
  year={2017},
  publisher={IEEE}
}

@inproceedings{czempiel2020tecno,
  title={Tecno: Surgical phase recognition with multi-stage temporal convolutional networks},
  author={Czempiel, Tobias and Paschali, Magdalini and Keicher, Matthias and Simson, Walter and Feussner, Hubertus and Kim, Seong Tae and Navab, Nassir},
  booktitle={International conference on medical image computing and computer-assisted intervention},
  pages={343--352},
  year={2020},
  organization={Springer}
}

@article{jin2021temporal,
  title={Temporal memory relation network for workflow recognition from surgical video},
  author={Jin, Yueming and Long, Yonghao and Chen, Cheng and Zhao, Zixu and Dou, Qi and Heng, Pheng-Ann},
  journal={IEEE Transactions on Medical Imaging},
  volume={40},
  number={7},
  pages={1911--1923},
  year={2021},
  publisher={IEEE}
}

@inproceedings{gao2021trans,
  title={Trans-svnet: Accurate phase recognition from surgical videos via hybrid embedding aggregation transformer},
  author={Gao, Xiaojie and Jin, Yueming and Long, Yonghao and Dou, Qi and Heng, Pheng-Ann},
  booktitle={International conference on medical image computing and computer-assisted intervention},
  pages={593--603},
  year={2021},
  organization={Springer}
}

@inproceedings{liu2023skit,
  title={Skit: a fast key information video transformer for online surgical phase recognition},
  author={Liu, Yang and Huo, Jiayu and Peng, Jingjing and Sparks, Rachel and Dasgupta, Prokar and Granados, Alejandro and Ourselin, Sebastien},
  booktitle={Proceedings of the IEEE/CVF international conference on computer vision},
  pages={21074--21084},
  year={2023}
}

@article{liu2025lovit,
  title={Lovit: Long video transformer for surgical phase recognition},
  author={Liu, Yang and Boels, Maxence and Garcia-Peraza-Herrera, Luis C and Vercauteren, Tom and Dasgupta, Prokar and Granados, Alejandro and Ourselin, Sebastien},
  journal={Medical Image Analysis},
  volume={99},
  pages={103366},
  year={2025},
  publisher={Elsevier}
}

@article{zhang2024adaptive,
  title={Adaptive graph learning from spatial information for surgical workflow anticipation},
  author={Zhang, Francis Xiatian and Deng, Jingjing and Lieck, Robert and Shum, Hubert PH},
  journal={IEEE Transactions on Medical Robotics and Bionics},
  year={2024},
  publisher={IEEE}
}

@article{yuan2022anticipation,
  title={Anticipation for surgical workflow through instrument interaction and recognized signals},
  author={Yuan, Kun and Holden, Matthew and Gao, Shijian and Lee, Wonsook},
  journal={Medical Image Analysis},
  volume={82},
  pages={102611},
  year={2022},
  publisher={Elsevier}
}

@inproceedings{silva2022analysis,
  title={Analysis of current deep learning networks for semantic segmentation of anatomical structures in laparoscopic surgery},
  author={Silva, Bruno and Oliveira, Bruno and Morais, Pedro and Buschle, LR and Correia--Pinto, Jorge and Lima, Estev{\~a}o and Vila{\c{c}}a, Joao L},
  booktitle={2022 44th annual international conference of the IEEE engineering in medicine \& biology society (EMBC)},
  pages={3502--3505},
  year={2022},
  organization={IEEE}
}

@inproceedings{rivoir2020rethinking,
  title={Rethinking anticipation tasks: Uncertainty-aware anticipation of sparse surgical instrument usage for context-aware assistance},
  author={Rivoir, Dominik and Bodenstedt, Sebastian and Funke, Isabel and von Bechtolsheim, Felix and Distler, Marius and Weitz, J{\"u}rgen and Speidel, Stefanie},
  booktitle={International conference on medical image computing and computer-assisted intervention},
  pages={752--762},
  year={2020},
  organization={Springer}
}

@article{bolya2025perception,
  title={Perception encoder: The best visual embeddings are not at the output of the network},
  author={Bolya, Daniel and Huang, Po-Yao and Sun, Peize and Cho, Jang Hyun and Madotto, Andrea and Wei, Chen and Ma, Tengyu and Zhi, Jiale and Rajasegaran, Jathushan and Rasheed, Hanoona and others},
  journal={arXiv preprint arXiv:2504.13181},
  year={2025}
}

@article{wu2023clipself,
  title={Clipself: Vision transformer distills itself for open-vocabulary dense prediction},
  author={Wu, Size and Zhang, Wenwei and Xu, Lumin and Jin, Sheng and Li, Xiangtai and Liu, Wentao and Loy, Chen Change},
  journal={arXiv preprint arXiv:2310.01403},
  year={2023}
}

@article{karypidis2024dino,
  title={Dino-foresight: Looking into the future with dino},
  author={Karypidis, Efstathios and Kakogeorgiou, Ioannis and Gidaris, Spyros and Komodakis, Nikos},
  journal={arXiv preprint arXiv:2412.11673},
  year={2024}
}

@article{xi2026grounding,
  title={Grounding Radiology Report Findings into Medical Image Segmentation},
  author={Xi, Suyang and Hu, Songtao and Wang, Shansong and Ding, Hong and Li, Yuheng and He, Wenfeng and Zhang, Kuo and Del Balzo, Luke and Zhong, Chong and Hu, Mingzhe and others},
  journal={npj Digital Medicine},
  year={2026},
  publisher={Nature Publishing Group UK London}
}

@article{xi2026medlvr,
  title={MedLVR: Latent Visual Reasoning for Reliable Medical Visual Question Answering},
  author={Xi, Suyang and Hu, Songtao and Lai, Yuxiang and Dan, Wangyun and Liu, Yaqi and Wang, Shansong and Yang, Xiaofeng},
  journal={arXiv preprint arXiv:2604.09757},
  year={2026}
}

@article{xi2026radiographic,
  title={A radiographic world model for clinical reasoning and evidence generation},
  author={Xi, Suyang and Hu, Songtao and Wang, Shansong and Safari, Mojtaba and del Balzo, Luke and Karim, Ehsan Ul and Hu, Mingzhe and Zhang, Kuo and Wang, Tonghe and Weichselbaum, Ralph R and others},
  journal={arXiv preprint arXiv:2609.07719},
  year={2026}
}

@article{xi2025uncertainty,
  title={Uncertainty as a foundation for trustworthy medical imaging ai: A comprehensive review},
  author={Xi, Suyang and Wang, Shansong and Safari, Mojtaba and Hu, Mingzhe and Tian, Zhen and Yang, Xiaofeng},
  year={2025},
  publisher={TechRxiv}
}

@inproceedings{yang2025enhancing,
  title={Enhancing Pathology Vision-Language Pre-Training: A Robust Baseline},
  author={Yang, Chenxi and Liu, Yunhao and Hu, Songtao and Liu, Tao and Ding, Hong and Xi, Suyang},
  booktitle={2025 IEEE International Conference on Bioinformatics and Biomedicine (BIBM)},
  pages={4931--4938},
  year={2025},
  organization={IEEE},
  doi={10.1109/BIBM66473.2025.11356606}
}
\end{document}